# Robust Trajectory Prediction Under Sensing and V2X Noise: Evaluation and Mitigation Strategies

Shahab Salehi, *Student Member, IEEE*, Luca Lusvarghi, *Member, IEEE*, Miguel Sepulcre, *Senior Member, IEEE*, and Javier Gozalvez, *Fellow, IEEE*

***Abstract*—Anticipating the future behavior of surrounding objects is key for the safe and efficient navigation of autonomous vehicles. State-of-the-art Transformer-based interaction-aware trajectory prediction models are typically trained and evaluated on large-scale, highly curated datasets obtained by filtering any object detection and tracking noise through extensive offline post-processing. However, in real-world deployments, the detected objects' state (position, heading, and speed) estimates are inherently affected by sensing and localization uncertainties. These uncertainties can be even higher for connected and automated vehicles receiving information from other road users through Vehicle-to-Everything (V2X) communications. In this paper, we first investigate the sensitivity of five state-of-the-art Transformer-based interaction-aware trajectory prediction models to realistic noisy object state estimates. Numerical results show that prediction accuracy degrades by up to a 6x factor as the noise level increases, highlighting the models' sensitivity to the mismatch between the noise conditions encountered during training and those experienced at deployment. We then show that retraining models on a more realistic noise-augmented dataset is a more effective strategy to mitigate the impact of noisy object state estimates on trajectory prediction accuracy than using denoising modules at the input. Retrained models limit the prediction accuracy degradation below a 1.5x factor. The noise-augmented dataset is openly released to support further research.**

## I. Introduction

AUTONOMOUS vehicles continuously monitor and anticipate the behavior of surrounding objects, such as pedestrians and other vehicles, to improve the safety and efficiency of their planned maneuvers. They rely on trajectory prediction models to estimate a target object's future state (i.e., its position, heading, and speed) over a given prediction window based on a sequence of observations of its past behavior. State-of-the-art trajectory prediction models, such as DeMo++ [1], SEPT [2], FutureNet-LOF [3], and QCNet [4], use powerful AI architectures, primarily based on Transformers, to learn the objects' driving behavior. By jointly modeling interactions among multiple surrounding objects (also known as multi-object interactions) and the relationships between the objects' mobility and the underlying road topology, these models can capture complex behavioral patterns and achieve higher trajectory prediction accuracy than conventional approaches based on kinematic models and state estimation techniques (e.g., Kalman filters).

State-of-the-art trajectory prediction models are commonly trained and evaluated on large-scale mobility datasets, such as the Argoverse 2 Motion Forecasting dataset[1] [5] and the INTERACTION dataset [6]. These datasets are collected using fleets of vehicles or drones and are carefully curated through extensive offline post-processing to reduce object detection and tracking errors. As a result, they provide accurate state (position, heading, and speed) estimates for a large number of objects, including objects that may be outside the sensing range of an individual vehicle. However, the state estimates obtained from onboard sensors are typically affected by sensing noise, which generally increases with the sensing distance and can result from hardware and software limitations (e.g., limited sensor resolution) or adverse environmental conditions (e.g., poor illumination and weather). Furthermore, in real-world deployments, an ego-vehicle can obtain information about objects outside its onboard sensing range only if such information is provided by other vehicles or the roadside infrastructure through Vehicle-to-Everything (V2X) communications. V2X communications introduce additional sources of uncertainty. In particular, the position of an object reported by another vehicle is typically expressed relative to the transmitter's own position and, therefore, localization errors at the transmitter propagate to the object's position received by the ego-vehicle, introducing additional positioning noise. Hereafter, we will refer to this V2X-related localization noise as V2X noise.

Several studies have shown that the discrepancy between the high-quality data used to train trajectory prediction models and the noisy conditions encountered during real-world deployment can lead to a significant performance degradation. In particular, [7] showed that the trajectory prediction accuracy of Recurrent Neural Network (RNN)-based models trained on curated datasets decreases substantially when the target object's state is perturbed with synthetic Gaussian noise representing typical distance-dependent sensing uncertainties. Similarly, [8], [9], and [10] reported substantial performance degradation for trajectory

This work was supported by the European Union under the 2023 MSCA Postdoctoral Fellowship program (project no. 101153845). This work was also partially funded by MICIU/AEI/10.13039/501100011033 and ERFD/EU (PID2023-150308OB-I00), and Generalitat Valenciana (CIAICO/2024/167).

*(Corresponding author: Shahab Salehi.)*

The authors are with the Networked Systems Laboratory, Universidad Miguel Hernández, Elche, Spain. E-mail: {ssalehi, llusvarghi, msepulcre, j.gozalvez}@umh.es.

[1] In the remainder of this work, we will refer to the Argoverse 2 Motion Forecasting dataset as the Argoverse 2 dataset.

prediction models based on Convolutional Neural Networks (CNNs), Variational AutoEncoders (VAEs), and Graph Neural Networks (GNNs), respectively, when the target-object states were affected by sensing noise. However, the sensitivity of recent Transformer-based interaction-aware models to noisy state estimates remains largely unexplored, including their sensitivity to the additional uncertainties introduced by V2X noise.

This work studies the sensitivity of five state-of-the-art Transformer-based interaction-aware trajectory prediction models – DeMo [11], DeMo++ [1], QCNet [4], SIMPL [12], and Forecast-MAE [13] – to noisy object state estimates affected by both sensing and V2X noise. The models are trained on the Argoverse 2 dataset and evaluated under different noise levels. Our results show that all five models are relatively robust to the distance-dependent sensing noise commonly considered in the literature, but their prediction accuracy degrades substantially under V2X noise. Numerical results show that, under high V2X noise conditions, the trajectory prediction accuracy of the analyzed models can degrade by up to a 6x factor. We then investigate two different strategies to mitigate the impact of noisy object state estimates on the models' trajectory prediction accuracy: (i) denoising the objects' state estimates using a Kalman filter and (ii) retraining the models on a more realistic noise-augmented version of the Argoverse 2 dataset. Denoising modules have been proposed in previous studies ([9], [10]) to mitigate the impact of noisy object state estimates. The obtained results show that retraining the models on noise-augmented data achieves better prediction accuracy under noisy inputs than the Kalman-filter-based denoising strategy evaluated in this work. Retrained models are more robust to all levels of sensing and V2X noise included in the training dataset, exhibiting a maximum degradation of 1.2x in trajectory prediction accuracy. We also tested both strategies under V2X noise levels beyond those encountered during training, with retrained models again achieving the best performance, with a maximum degradation of 1.5x in trajectory prediction accuracy. In contrast, Kalman filtering only partially mitigates the impact of noisy object state estimates: trajectory prediction accuracy decreases monotonically as the V2X noise level increases, reaching up to a 4x degradation. These results highlight the importance of training and evaluating trajectory prediction models with data that reflects real-world operating conditions, shedding light on the potential of state-of-the-art Transformer-based interaction-aware models to handle noisy input data. Exposure to realistic sensing and V2X noise levels enables these models to learn latent representations of driving behavior that are more robust to object state estimation uncertainties, while also improving their ability to handle noise levels beyond those encountered during training. To facilitate further research, we publicly release the noise-augmented version of the Argoverse 2 dataset employed in this work at [14].

The remainder of this paper is organized as follows. Section II reviews the evolution of trajectory prediction models and large-scale mobility datasets, as well as state-of-the-art approaches for mitigating the impact of noisy data on trajectory prediction. Next, Section III describes the experimental setup employed in this work. Section IV analyzes the sensitivity of Transformer-based interaction-aware trajectory prediction models to sensing and V2X noise, while Section V compares the two mitigation strategies considered in this work. Finally, Section VI draws the conclusions.

## II. Related Work

Trajectory prediction models can be broadly classified into physics-based and data-driven models. Physics-based models such as Kalman filters and Monte Carlo methods rely on kinematic assumptions to estimate future trajectories. Data-driven models rely on deep neural networks to learn motion patterns and interactions from large-scale mobility datasets. Early data-driven models were primarily based on CNNs and RNNs. Recent state-of-the-art models predominantly use Transformer-based architectures, which employ attention mechanisms to model temporal dependencies and interactions among multiple objects while incorporating information about the surrounding road topology. These Transformer-based models mainly differ in how they represent the driving scene, model interactions, and generate future trajectories. MTR [15] predicts multiple possible future behaviors and progressively refines their corresponding trajectories, while MTR++ [16] extends this approach to jointly predict the trajectories of multiple interacting objects. QCNet [4] uses a two-stage approach that first generates candidate trajectories and then refines them using surrounding traffic information, while SIMPL [12] efficiently captures interactions among surrounding objects and generates smooth trajectories using Bézier curves. DeMo [11] models possible movement directions and the evolution of an object's state over time separately, and then combines them to generate diverse and temporally consistent trajectories, while DeMo++ [1] extends this approach by incorporating information from previous predictions and refining the generated trajectories to improve accuracy and consistency over time. FutureNet-LOF [3] uses predicted future movements to capture relevant surrounding information and jointly predicts trajectories and lane occupancy. Recent studies have also explored different training strategies and trajectory prediction approaches. Forecast-MAE [13] and SEPT [2] use self-supervised pretraining to learn driving patterns by reconstructing missing information from trajectories and road maps, while SEPT further learns interactions between objects and roads during pretraining. SEAM [17] reuses information from previous predictions to generate new trajectories efficiently in real time, while Polaris [18] represents object positions using distances and angles instead of conventional Cartesian coordinates to better capture spatial relationships. Despite their differences in architecture and trajectory generation, these models generally rely on historical object states and road-topology information as inputs and predict multiple candidate future trajectories for each target object. Their training and evaluation are typically performed using large-scale, highly curated mobility datasets.

Existing datasets include both synthetic datasets generated using simulators such as SUMO or CARLA and real-world datasets collected from vehicles, drones, and roadside infrastructure. Synthetic datasets enable controlled and reproducible experimentation but often rely on simplified traffic models that cannot fully capture the complexity of real-world driving. Real-world datasets provide more representative mobility behavior and sensing environments. Early real-world datasets, such as nuScenes [19], were collected using the onboard sensors of a single vehicle, inherently limiting the spatial coverage and the number of recorded objects due to the sensing range of the recording vehicle's sensors. More recent large-scale datasets overcome these limitations by aggregating data collected from vehicle fleets, drones, and roadside infrastructure, providing information about a larger number of surrounding objects and offering richer contextual information for trajectory prediction. Representative examples of these large-scale datasets include the Argoverse 1 and Argoverse 2 Motion Forecasting datasets [5], the Waymo Open Motion Dataset [20], and the Lyft [21] and INTERACTION [6] datasets. These large-scale datasets generally rely on extensive offline post-processing to minimize object detection and tracking errors and to provide accurate object state information for trajectory prediction. Argoverse 2 Motion Forecasting is one of the largest and most widely used datasets for training and evaluating state-of-the-art trajectory prediction models. It includes 250,000 driving scenes covering 2,220 km of unique roadways across six cities. Its extensive spatial coverage offers the geographic diversity and depth required to train and evaluate state-of-the-art trajectory prediction models. However, the highly curated object states available in these datasets do not necessarily reflect the uncertainties affecting object state estimates during real-world operation.

Multiple existing studies have investigated the sensitivity of trajectory prediction models to noisy object state estimates resulting from sensing uncertainties. The study in [7] showed that adding zero-mean Gaussian noise with a standard deviation of up to 0.5 m to the objects' positions significantly reduced the prediction accuracy of an LSTM-based trajectory prediction model. Similarly, the authors of [8] analyzed the sensitivity of both CNN- and LSTM-based models by perturbing only the heading of the detected objects with a 90° offset, reporting a substantial degradation in trajectory prediction accuracy. Studies on pedestrian trajectory prediction have explored more advanced AI architectures, including Conditional VAEs (CVAEs) and GNNs. For example, [10] evaluated trajectory prediction models under Gaussian noise applied to historical trajectories and demonstrated that noisy observations can significantly degrade prediction accuracy. In [9], the authors evaluated several trajectory prediction models under position noise with standard deviations ranging from 0.2 m to 0.4 m and reported a consistent degradation in prediction accuracy as the noise intensity increased. Several strategies have been proposed to reduce the impact of noisy object state estimates. One option is to denoise the input states before they are provided to the trajectory prediction model. Such denoising can be performed using Kalman filters [22] or dedicated AI-based denoising modules [23]. Another approach is to design trajectory prediction models able to natively cope with noisy input data. For example, [24] proposes an AI architecture for pedestrian trajectory prediction that simultaneously denoises historical observations and predicts future trajectories to improve robustness. In [7], the authors propose an LSTM-based architecture using polar-coordinate representations to improve robustness to positional noise and evaluate it under noise with a standard deviation of 0.5 m. Overall, existing studies provide evidence that noisy object state estimates can degrade trajectory prediction accuracy and have explored both input denoising strategies and custom designs to mitigate this degradation. However, the sensitivity of recent Transformer-based interaction-aware models to realistic sensing noise remains insufficiently studied. The impact of the additional positioning uncertainties that can arise when object information is received through V2X communications has not yet been investigated.

## III. Experimental Setup

We analyze the sensitivity of Transformer-based interaction-aware trajectory prediction models to sensing and V2X noise using five state-of-the-art models trained on the Argoverse 2 dataset: SIMPL, Forecast-MAE, QCNet, DeMo, and DeMo++. At the time of writing, these models rank 134th, 80th, 13th, 6th, and 8th, respectively, on the Argoverse 2 Motion Forecasting leaderboard [25]. We selected these models because pre-trained weights for their leaderboard-ranked versions are publicly available.

### *A. Trajectory Prediction Models*

All evaluated trajectory prediction models use the same types of input and output data. The input consists of the temporal evolution of the object's state (position, heading, and speed) over a tracking window of $T_{\mathrm{H}}$ seconds (i.e., the object history) and information about the road topology. These inputs are represented in vectorized form, using sequences of points and polylines to describe object histories and road topology. The models' output consists of a multimodal prediction of six candidate future trajectories stretching over a prediction window of $T_{\mathrm{F}}$ seconds for each target object, with each candidate trajectory associated with a confidence score.

The evaluated models mainly differ in how they model interactions between objects and generate the predicted trajectories. DeMo [11] introduces a motion-decoupling strategy that separately models the target object's high-level driving intention (e.g., going straight or turning left) and fine-grained spatio-temporal motion, before combining the two through attention mechanisms. DeMo++ [1] extends DeMo with a cross-scene intention interaction module and a trajectory refinement stage that improves spatial consistency. QCNet [4] adopts a query-centric encoding framework in which objects and road topology are represented in local coordinate systems, enabling efficient feature reuse across time. Trajectories are generated through a two-stage decoder consisting of an initial proposal stage followed by a refinement stage. SIMPL [12] focuses on computational efficiency and inference speed by jointly predicting the trajectories of all traffic participants in a single forward pass using a Symmetric Fusion Transformer. Instead of predicting discrete positions, it estimates Bézier control points to produce smooth and physically consistent trajectories. Forecast-MAE [13] differs from the other models by incorporating a self-supervised pretraining stage based on a Masked AutoEncoder (MAE). During pretraining, it reconstructs masked object trajectories and map elements to learn general spatiotemporal representations before being fine-tuned for trajectory prediction.

*B. Argoverse 2*

Argoverse 2 [5] is a large-scale dataset employed for the training, evaluation, and benchmarking of trajectory prediction models. It contains approximately 250,000 driving scenes recorded across six U.S. cities (Austin, Detroit, Miami, Pittsburgh, Palo Alto, and Washington, D.C.) using a fleet of vehicles equipped with two rooftop LiDAR sensors, seven ring cameras, and two front-facing stereo cameras. Each driving scene is $T$ = 11 seconds long and includes annotated trajectories for the ego-vehicle and all other surrounding objects. The ego-vehicle is one of the vehicles in the recording fleet, while the other recording vehicles are encoded as objects. Objects are classified into 10 distinct categories (e.g., vehicles, pedestrians, cyclists) and their trajectories are divided into a past observation window of $T_{\mathrm{H}}$ = 5 seconds (the object's history) and a future prediction horizon of $T_{\mathrm{F}}$ = 6 seconds. Each trajectory represents the evolution of the object's state (position, speed, and heading) and is sampled at 10 Hz. To obtain smooth and continuous trajectories, the objects' states are estimated at each timestep via a combination of GPS-based localization, onboard sensing, and offline sensor fusion and post-processing techniques.

A distinctive feature of Argoverse 2 is its extended spatial coverage and the availability of rich contextual information, obtained by processing offline observations collected by multiple vehicles throughout each driving scene. This approach allows the dataset to include objects located up to 250 meters from the ego-vehicle. Such extended spatial coverage provides trajectory prediction models with information about a larger set of surrounding objects, which can help them capture complex multi-object interactions and improve trajectory prediction accuracy. However, in real-world deployments, autonomous vehicles can obtain information about distant or occluded objects beyond their onboard sensing range only through V2X communications. The characteristics and impact of information received through V2X communications are not explicitly modeled in Argoverse 2.

*C. Noise Models*

Incorporating realistic uncertainties into existing datasets is essential for training and evaluating state-of-the-art trajectory prediction models. To this end, we perturb the object states in the Argoverse 2 dataset using two different types of noise that represent common real-world uncertainties in object state estimation: sensing and V2X noise.

At each timestep, the position, speed, and heading of objects locally detected by the ego-vehicle through its onboard sensors are perturbed by zero-mean Gaussian noise $N_{\mathrm{S}} \sim \mathcal{N}(0, \sigma_{\mathrm{S}}^2)$. This type of noise captures local perception uncertainties arising from adverse environmental conditions (e.g., poor weather and illumination) as well as hardware and software limitations of onboard sensors. Following [26], the intensity of the sensing noise depends on the distance between the detected object, $o_i$, and the ego-vehicle. Specifically, its standard deviation depends linearly on distance and is defined as $\sigma_{\mathrm{S}} = 0.001 \times d_i$, where $d_i$ denotes the distance between $o_i$ and the ego-vehicle. Sensing noise is directly applied to the objects' position in meters (m), while the corresponding speed and heading perturbations follow the same distance-dependent law and are expressed in their respective units (m/s and radians).

We model the probability that an object $o_i$ can be locally detected by the ego-vehicle using its onboard sensors through a detection probability $P(d_i)$. If another object $o_j$ blocks the direct line of sight between the ego-vehicle and $o_i$, the target is completely occluded, yielding a detection probability of $P(d_i) = 0$. Otherwise, in the absence of physical occlusion, the detection probability depends on the distance $d_i$ between $o_i$ and the ego-vehicle, as follows:

$$P(d_i) = \begin{cases} 1 & d_i \leq 90 \\ \dfrac{140 - d_i}{140 - 90} & 90 < d_i < 140 \\ 0 & d_i \geq 140 \end{cases} \quad (1)$$

The piecewise detection probability in (1) is consistent with vehicular perception ranges considered in prior literature [27].

For objects outside the ego-vehicle's sensing range, information can instead be received through V2X communications from other vehicles or roadside infrastructure. The position of an object reported by a transmitting vehicle is typically determined relative to the transmitter's own position. Consequently, errors in the transmitter's localization affect the reported position of the object and propagate to the position estimate received by the ego-vehicle. We model this additional positioning uncertainty, referred to as V2X noise, as a zero-mean Gaussian noise source, $N_{V2X} \sim \mathcal{N}(0, \sigma_{V2X}^2)$ [28], whose standard deviation $\sigma_{V2X}$ represents the magnitude of the transmitter-side localization errors. In this work, we consider V2X noise levels up to $\sigma_{V2X}$ = 2.0 m, consistent with the GNSS noise model used in [26], and further test higher noise levels of up to $\sigma_{V2X}$ = 5.0 m to evaluate the proposed solution under more challenging conditions.[2]

## IV. Noise Sensitivity

This section investigates the sensitivity of state-of-the-art trajectory prediction models to object-state estimation uncertainties arising from sensing and V2X noise. We first examine DeMo++ in detail as it ranks among the top ten models on the Argoverse 2 Motion Forecasting leaderboard, and is the most recent of the models considered in this study. We analyze the impact of increasing noise levels on its qualitative predictions and quantitative performance across the Argoverse 2 dataset. We then compare its sensitivity with that of four other trajectory prediction models to assess whether the observed performance degradation is specific to DeMo++ or represents a broader limitation of current interaction-aware prediction approaches.

[2] In this work, we intentionally focus on noise sources arising from perception and localization uncertainties and assume ideal V2X communications. Therefore, we do not consider additional noise contributions associated with V2X communication reliability or latency.

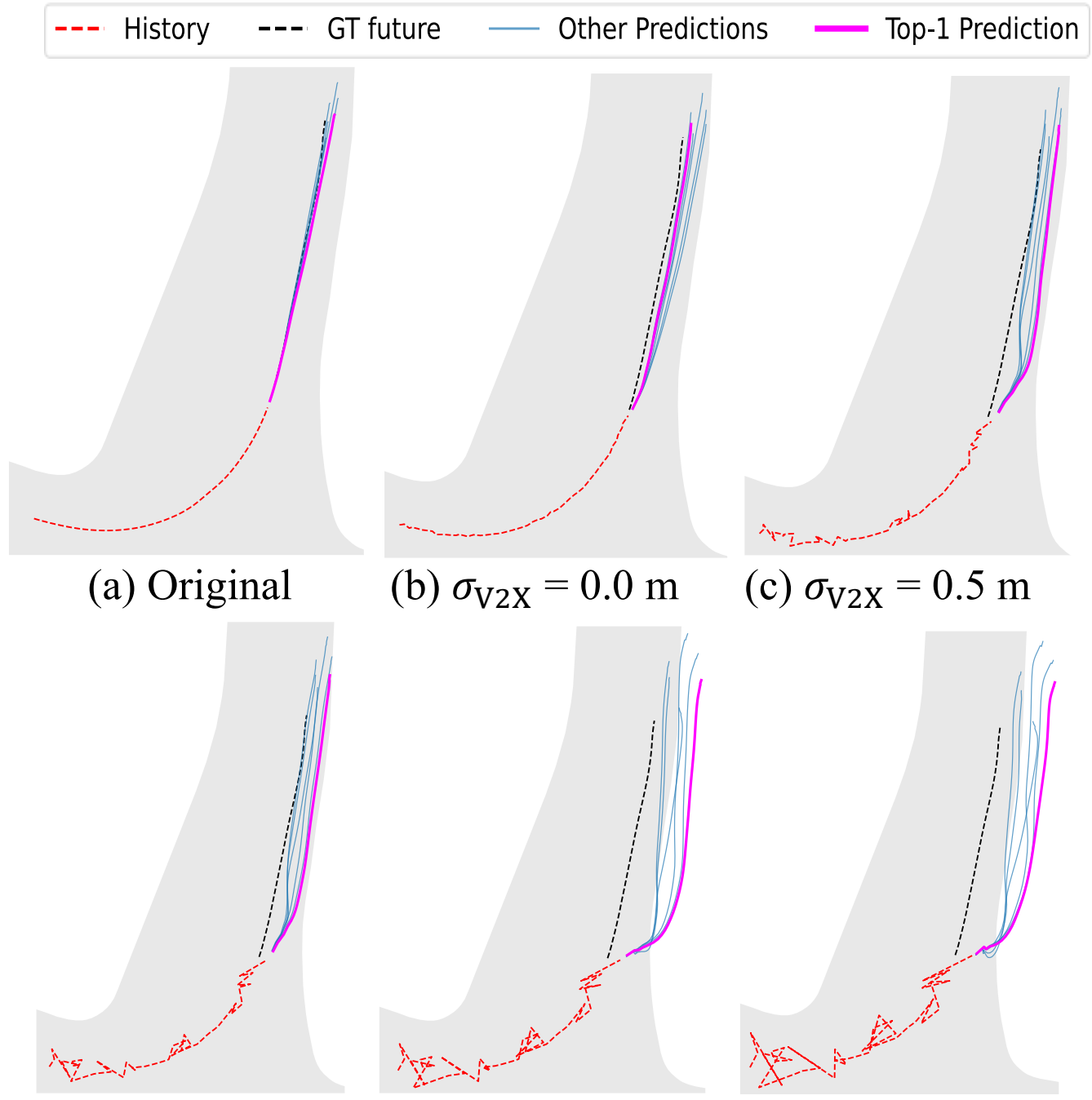

(a) Original (b) $\sigma_{\mathrm{V2X}} = 0.0$ m (c) $\sigma_{\mathrm{V2X}} = 0.5$ m

(d) $\sigma_{\mathrm{V2X}} = 1.0$ m (e) $\sigma_{\mathrm{V2X}} = 1.5$ m (f) $\sigma_{\mathrm{V2X}} = 2.0$ m

**Fig. 1.** DeMo++ trajectory predictions under different levels of object-state estimation noise.

### *A. DeMo++*

We begin by analyzing the sensitivity of DeMo++ to object state estimation uncertainties. Fig. 1(a)–(f) qualitatively illustrate the impact of noisy object state estimates on DeMo++ trajectory prediction accuracy in a representative Argoverse 2 driving scene. Each figure reports the target object's past trajectory (or history) over the last $T_H$ seconds, its Ground-Truth (GT) future trajectory over the next $T_F$ seconds, and the six trajectories predicted by DeMo++. The predicted trajectory with the highest confidence score is denoted as the top-1 prediction. Fig. 1(a) shows DeMo++ predictions in the original high-quality Argoverse 2 driving scene, where the objects' states are not perturbed by either sensing noise ($\sigma_S = 0.0$ m) or V2X noise ($\sigma_{V2X} = 0.0$ m). In this setting, all six predicted trajectories closely align with the GT future trajectory. Fig. 1(b)–1(f) show the predictions when the histories of the target object and surrounding objects are perturbed by sensing noise, following the model described in Section III-C, and increasing levels of V2X noise. The V2X noise standard deviation ranges from $\sigma_{V2X} = 0.0$ m (i.e., there is only sensing noise) to $\sigma_{V2X} = 2.0$ m. The results reveal a progressive degradation in prediction accuracy as V2X noise increases. With only sensing noise (Fig. 1(b)), DeMo++ predicted trajectories are nearly identical to those obtained in the noise-free scenario (Fig. 1(a)). This is also the case under low V2X noise conditions (Fig. 1(c), $\sigma_{V2X} = 0.5$ m). However, at intermediate V2X noise levels ($\sigma_{V2X} = 1.0$ m, Fig. 1(d)), the predicted trajectories begin to noticeably diverge from the ground-truth reference. At this noise level, V2X noise becomes the largest source of uncertainty as the maximum standard deviation of sensing noise is $\sigma_S = 0.14$ m at the maximum object-ego-vehicle distance. The degradation of the trajectory prediction accuracy becomes particularly severe under high V2X noise levels (Fig. 1(e) and 1(f), $\sigma_{V2X} \geq 1.5$ m), with predicted trajectories substantially deviating from the GT reference, even extending beyond the road boundaries. These results suggest that DeMo++ has limited robustness to object-state estimation errors not represented in its training data. Furthermore, under high-noise conditions, the top-1 prediction is no longer the most accurate of the six predicted trajectories, indicating that noise also compromises the model's ability to rank its predictions reliably.

To complement the qualitative insights from Fig. 1, we quantify the sensitivity of DeMo++ to object state estimation uncertainties across the entire Argoverse 2 dataset using three metrics commonly employed in trajectory prediction studies: Minimum Final Displacement Error ($\mathrm{minFDE}_K$), Minimum Average Displacement Error ($\mathrm{minADE}_K$), and Miss Rate ($\mathrm{MR}_K$). The $\mathrm{minFDE}_K$ metric measures the Euclidean distance between the final point of the best predicted trajectory and the final point of the GT trajectory. Here, the best trajectory is the one among the $K$ predicted trajectories whose final position is closest to the ground-truth final position. The $\mathrm{minFDE}_K$ metric can be mathematically expressed as:

$$\mathrm{minFDE}_K = \min_{k \in \{1,\dots,K\}} \left\| p_{\mathrm{final}}^{k} - p_{\mathrm{final}}^{\mathrm{gt}} \right\|_2, \tag{2}$$

where $p_{\mathrm{final}}^{k}$ and $p_{\mathrm{final}}^{\mathrm{gt}}$ indicate the (x, y) coordinates of the final point of the $k$-th predicted trajectory and the ground-truth trajectory, respectively. The $\mathrm{minADE}_K$ metric measures the average Euclidean distance between all points of the best predicted trajectory and the ground-truth trajectory. Let $N$ indicate the total number of points in each predicted trajectory and $p_n^{\mathrm{best}}$ and $p_n^{gt}$ denote the $n$-th ($n = 1, \dots, N$) pair of coordinates in the best and ground truth trajectory, respectively. $\mathrm{minADE}_K$ can be mathematically expressed as:

$$\mathrm{minADE}_K = \frac{1}{N} \sum_{n=1}^{N} \left\| p_n^{\mathrm{best}} - p_n^{\mathrm{gt}} \right\|_2. \tag{3}$$

The $\mathrm{MR}_K$ metric corresponds to the fraction of target objects for which the $\mathrm{minFDE}_K$ exceeds 2.0 m, meaning that all $K$ predicted trajectories have a final position more than 2.0 m away from the ground-truth final position. For each target object, a miss is defined as:

$$\mathrm{MR}_K = \begin{cases} 0, & \exists k \in \{1,\dots,K\}, \left\| p_{\mathrm{final}}^{k} - p_{\mathrm{final}}^{\mathrm{gt}} \right\|_2 \leq 2 \\ 1, & \text{otherwise} \end{cases}. \tag{4}$$

All three metrics ($\mathrm{minFDE}_K$, $\mathrm{minADE}_K$, and $\mathrm{MR}_K$) are averaged across the entire dataset and reported in Table I for $K = 1$ (Table I(a)) and $K = 6$ (Table I(b)) considering V2X noise levels from $\sigma_{\mathrm{V2X}} = 0.0$ m to $\sigma_{\mathrm{V2X}} = 2.0$ m. All results reported in Table I, except those for the original noise-free setting, account for both sensing and V2X noise. When $K = 1$, metrics are calculated considering only the top-1 predicted trajectory, which is the trajectory with the highest confidence score. When $K = 6$, metrics are calculated considering the best predicted trajectory among the six candidates provided by the model. Comparing the two tables reveals that the trajectory with the highest confidence score is typically not the most accurate one,

TABLE I
IMPACT OF NOISE ON DEMO++ PREDICTION ACCURACY

| $\sigma_{V2X}$ [m] | minFDE$_1$ [m] | minADE$_1$ [m] | MR$_1$ |
|---|---|---|---|
| No noise (original) | 3.70 | 1.50 | 0.55 |
| 0.0 | 4.20 | 1.77 | 0.60 |
| 0.1 | 4.43 | 1.91 | 0.62 |
| 0.2 | 4.67 | 2.06 | 0.64 |
| 0.5 | 5.50 | 2.64 | 0.67 |
| 1.0 | 6.90 | 3.54 | 0.69 |
| 1.5 | 8.10 | 4.25 | 0.70 |
| 2.0 | 9.03 | 4.79 | 0.71 |

(a) $K = 1$

| $\sigma_{V2X}$ [m] | minFDE$_6$ [m] | minADE$_6$ [m] | MR$_6$ |
|---|---|---|---|
| No noise (original) | 1.12 | 0.61 | 0.12 |
| 0.0 | 1.39 | 0.86 | 0.16 |
| 0.1 | 1.48 | 0.95 | 0.17 |
| 0.2 | 1.57 | 1.03 | 0.19 |
| 0.5 | 1.88 | 1.33 | 0.23 |
| 1.0 | 2.52 | 1.82 | 0.28 |
| 1.5 | 3.00 | 2.18 | 0.31 |
| 2.0 | 3.40 | 2.46 | 0.33 |

(b) $K = 6$

even at low V2X noise levels. In fact, the $K = 1$ metrics are consistently worse (i.e., larger) than their $K = 6$ counterparts. This result highlights the inherent challenges of trajectory prediction models in correctly ranking their predicted trajectories, a well-known challenge in multimodal trajectory prediction [29].

Table I clearly shows the high sensitivity of DeMo++ trajectory prediction accuracy to object state estimation uncertainties. All three metrics increase monotonically with V2X noise. In Table I(a), minADE$_1$ increases from 1.50 m in the original noise-free dataset to 4.79 m under high-noise conditions ($\sigma_{V2X} = 2.0$ m), representing more than a 3-fold increase. A similar degradation is observed for $K = 6$ (Table I(b)), where minADE$_6$ increases from 0.61 m to 2.46 m.

Fig. 2 complements Table I by illustrating the relative increase of the minFDE$_K$, minADE$_K$, and MR$_K$ metrics with respect to the original noise-free Argoverse 2 dataset for different V2X noise levels. As in Table I, all results except the noise-free baseline account for both sensing and V2X noise. Fig. 2(a) shows that minFDE$_1$ and minADE$_1$ increase by approximately 1.2x and 1.3x, respectively, even at very low noise levels ($\sigma_{V2X} = 0.1$ m) when $K = 1$. At $\sigma_{V2X} = 2.0$ m, the two metrics increase by 2.4x and 3.2x, respectively, with respect to their noise-free baseline values. The relative increase in MR$_1$ remains below 1.3x across all evaluated V2X noise levels. This smaller relative increase is explained by the already high baseline miss rate of 0.55 reported in Table I. The degradation is even more pronounced for some metrics when $K = 6$ (Fig. 2(b)), where the relative increase of minADE$_6$ can be as large as a 4.0x factor at $\sigma_{V2X} = 2.0$ m. Fig. 2(a) and 2(b) show that minADE$_K$ exhibits a consistently greater relative increase than

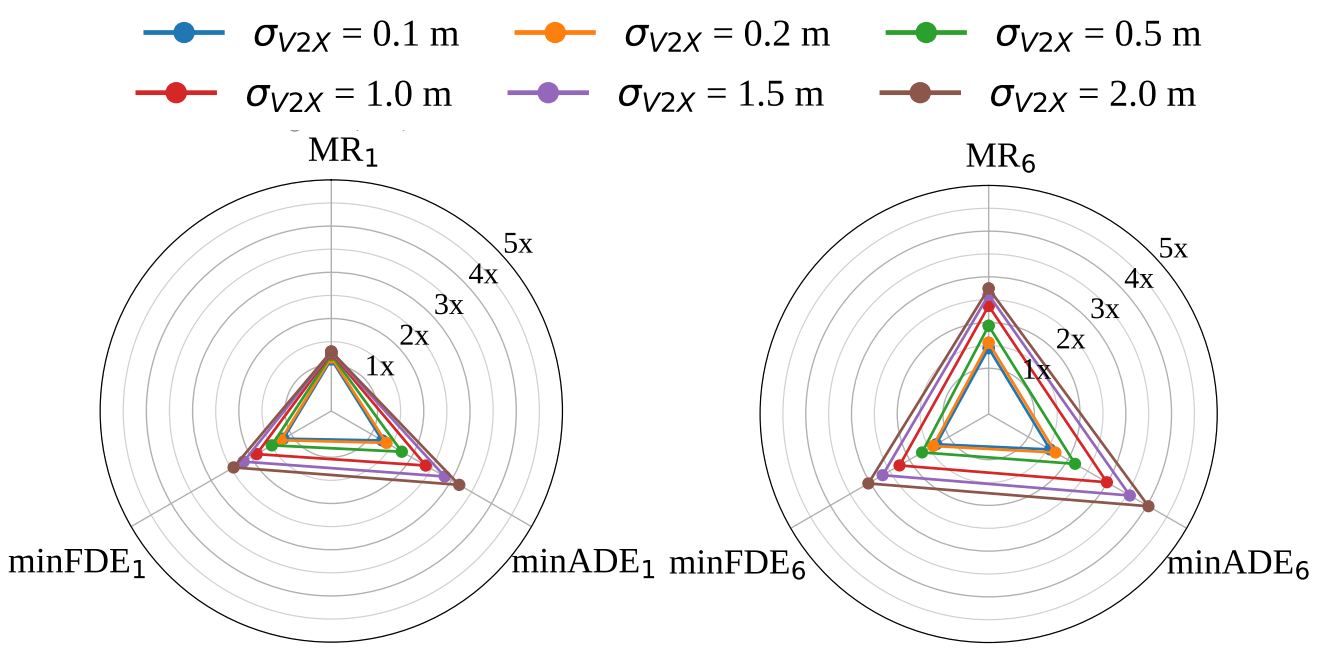

(a) $K = 1$ (b) $K = 6$

**Fig. 2.** Relative increase of the $minADE_K$, $minFDE_K$, and $MR_K$ metrics for DeMo++ with respect to the original Argoverse 2 noise-free scenario under increasing V2X noise levels.

minFDE$_K$. This indicates that object state estimation uncertainties have a larger impact on the overall shape of the predicted trajectories than on the final predicted positions.

Fig. 3(a)–(c) analyze the effect of the target object's visibility on the minFDE$_K$, minADE$_K$, and MR$_K$ metrics, respectively. The figures compare the metrics obtained using the original noise-free Argoverse 2 dataset with those obtained under high V2X noise ($\sigma_{V2X} = 2.0$ m). When the target object is not visible to the ego-vehicle's onboard sensors, its state information is obtained through V2X communications and is affected by V2X noise. Therefore, we investigate how different visibility levels influence trajectory prediction accuracy under V2X noise. The visibility level is defined as the percentage of the target object's history during which it is directly detected by the ego-vehicle's onboard sensors and its state estimates are therefore affected by sensing noise ($N_S$) only. The results show that the target

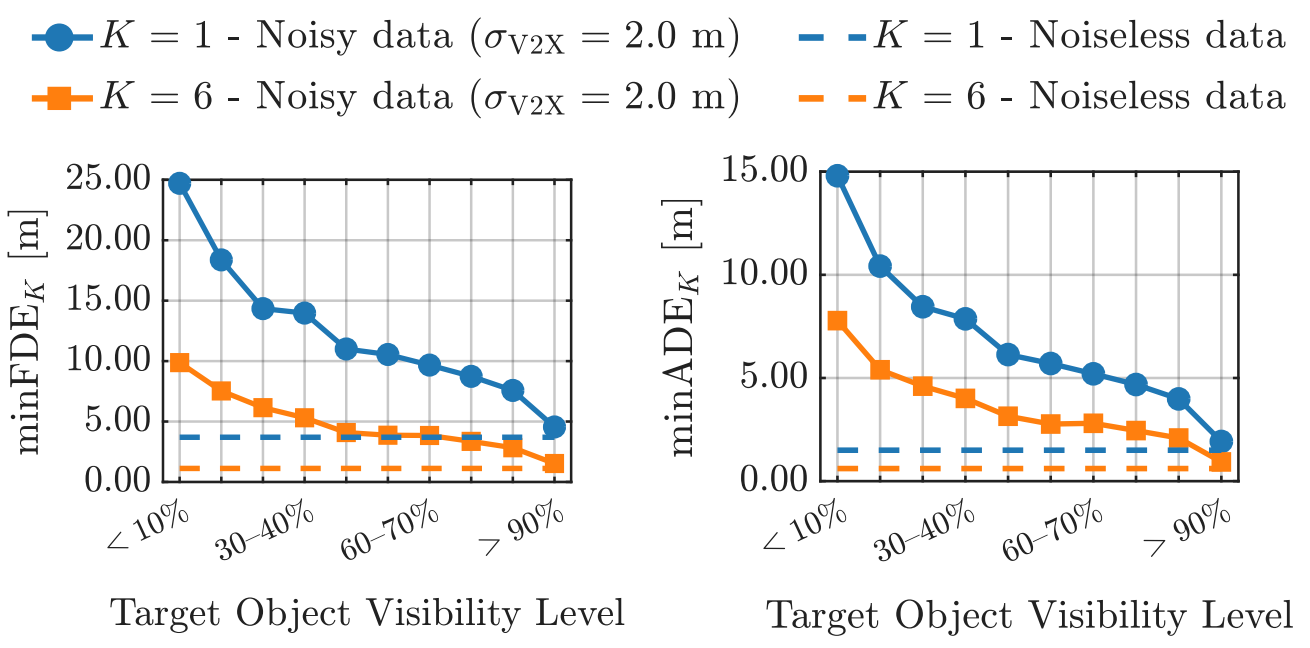

(a) minFDE$_K$ (b) minADE$_K$

(c) MR$_K$

**Fig. 3.** Impact of the target object's visibility on DeMo++ trajectory prediction accuracy under noise-free and noisy ($\sigma_{V2X} = 2.0$ m) scenarios.

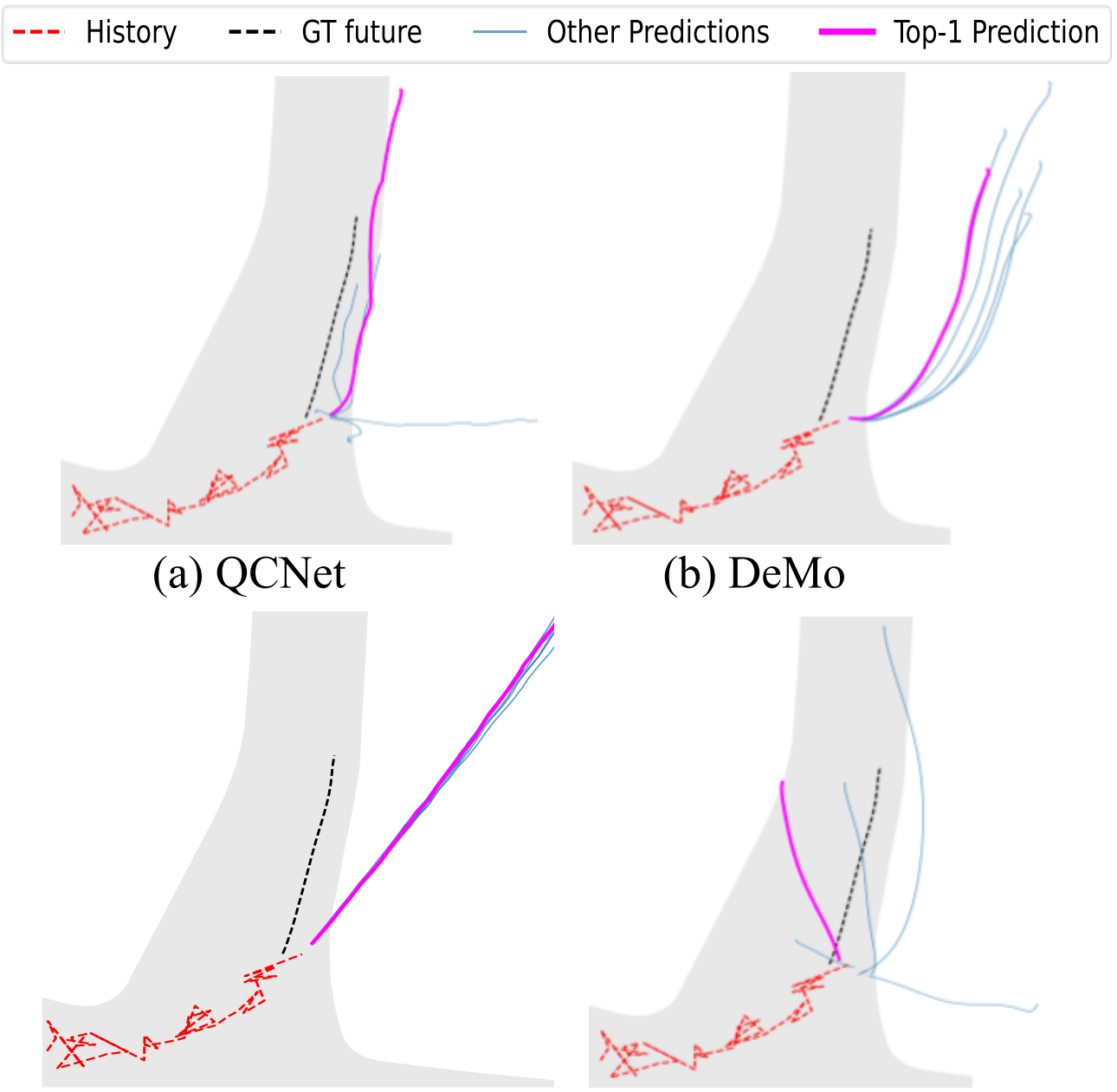

(a) QCNet (b) DeMo

(c) Forecast-MAE (d) SIMPL

**Fig. 4.** Qualitative comparison of trajectory predictions within a representative Argoverse 2 scenario under high noise (sensing noise and V2X noise with $\sigma_{V2X}$ = 2.0 m).

object's visibility level strongly influences DeMo++ performance and that prediction accuracy deteriorates even when visibility exceeds 90%. For visibility levels above 90%, minFDE$_6$, minADE$_6$, and MR$_6$ increase by approximately 37%, 52%, and 58%, respectively. This indicates that even a small proportion of observations affected by V2X noise can substantially reduce trajectory prediction accuracy. The degradation of trajectory prediction accuracy becomes increasingly severe as the visibility level decreases. All metrics increase sharply when visibility drops below 10%, as the model relies predominantly on V2X information rather than direct onboard observations. These results demonstrate that noisy V2X data can severely impair the accuracy of trajectory prediction models trained with noise-free data, particularly when direct observations of the target object are limited.

### *B. Comparative Analysis*

Fig. 4 extends the analysis to the other four trajectory prediction models considered in this work: QCNet, DeMo, Forecast-MAE, and SIMPL. For each model, the figure plots the target object's past trajectory, its ground-truth (GT) future trajectory, and the six predicted trajectories. The top-1 prediction refers to the predicted trajectory with the highest confidence score. The figure focuses on a high-noise scenario that includes both sensing and V2X noise, as described in Section III-C , with $\sigma_{V2X}$ = 2.0 m. As observed for DeMo++ in Fig. 1(f), all four models exhibit a substantial degradation in trajectory prediction accuracy under high noise levels. However, despite this common trend, the error patterns vary significantly across the compared models. QCNet (Fig. 4(a)) and SIMPL (Fig. 4(d)) tend to generate spatially dispersed predictions, including trajectories that extend beyond the road boundaries and deviate significantly from the GT trajectory. In contrast, DeMo++ (Fig. 1(f)), DeMo (Fig. 4(b)), and Forecast-MAE (Fig. 4(c)) produce less spatially dispersed predictions, generally clustered around a common forward direction that is largely consistent with the direction of the GT trajectory. These results indicate that the sensitivity of DeMo++ to object state estimation uncertainties is not specific to its architecture, but rather a common problem that applies to other Transformer-based interaction-aware trajectory prediction models trained on high-quality, noise-free datasets such as Argoverse 2.

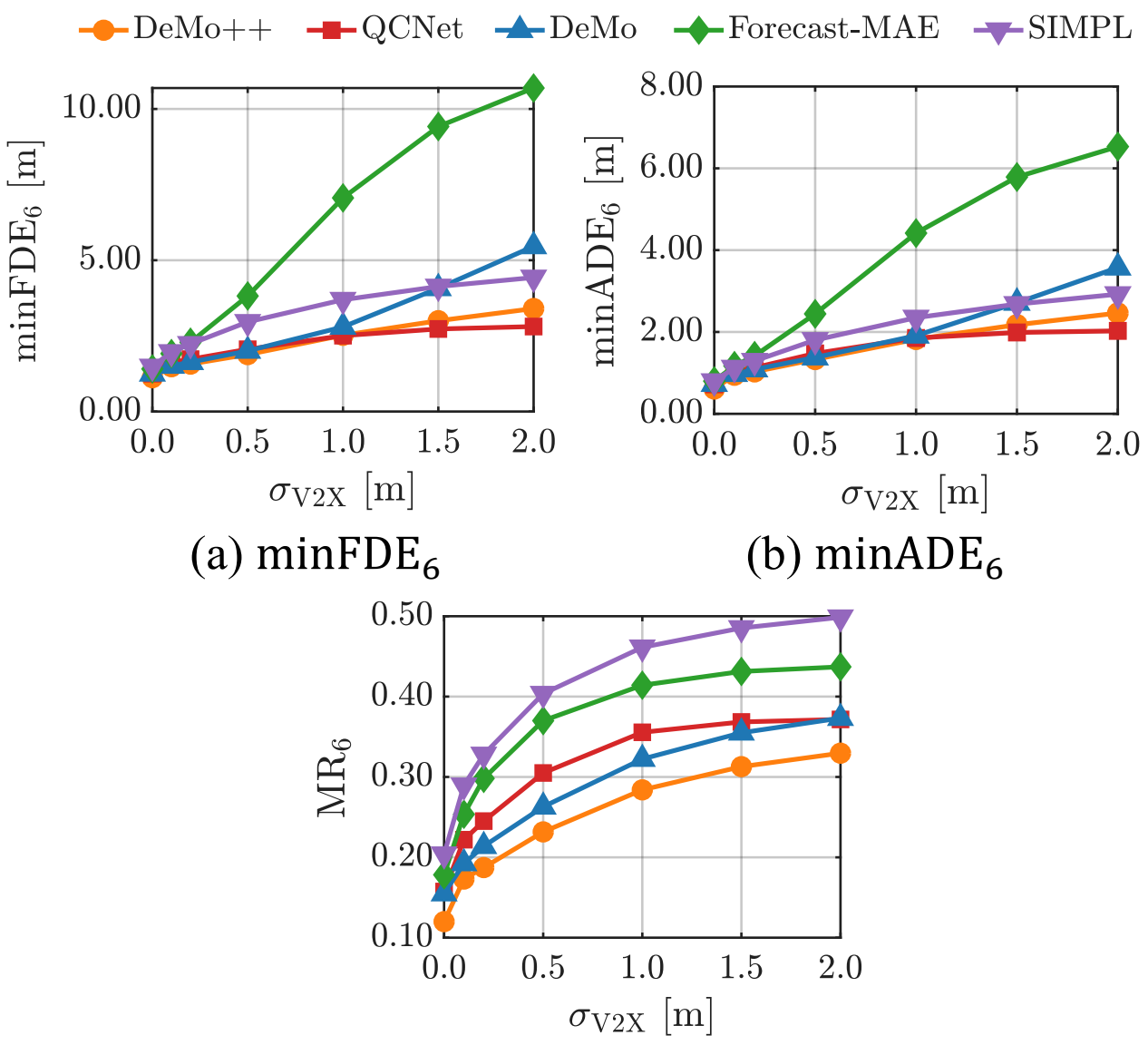

(a) minFDE$_6$ (b) minADE$_6$

(c) MR$_6$

**Fig. 5.** Impact of increasing V2X noise ($N_{V2X}$) on the performance of five evaluated models ($K$ = 6).

To confirm this observation, Fig. 5 quantitatively compares the impact of object state estimation uncertainties on the prediction accuracy of the five models. Fig. 5(a)-(c) report the minFDE$_6$, minADE$_6$, and MR$_6$ metrics, respectively. Since the results for $K$ = 1 and $K$ = 6 show similar trends, we focus on $K$ = 6 to avoid repetition. The results show that the prediction accuracy of all five models significantly deteriorates as the noise level increases, with notable differences in robustness among the evaluated models. In particular, the models ranked lower on the Argoverse 2 Motion Forecasting leaderboard are more sensitive to object state estimation uncertainties. Forecast-MAE (ranked 80th) exhibits the largest minFDE$_6$ and minADE$_6$ values, standing out as the most sensitive model among the evaluated ones. SIMPL (134th) also exhibits a strong sensitivity, as its minFDE$_6$ can be as large as 5 m when $\sigma_{V2X}$ = 2.0 m. In contrast, higher-ranked models achieve significantly smaller values across all metrics. For example, minFDE$_6$ remains below 4 m for both DeMo++ and QCNet. While QCNet and DeMo++ exhibit comparable minFDE$_6$ and minADE$_6$ values, DeMo++ proves to be the most robust model due to its substantially lower MR$_6$.

## V. Mitigation Strategies

Section IV showed that state-of-the-art Transformer-based interaction-aware trajectory prediction models trained on curated noise-free datasets such as Argoverse 2 are highly sensitive to object state estimation uncertainties. This section evaluates two strategies for mitigating the sensitivity to noise using the least sensitive (DeMo++) and the most sensitive (Forecast-MAE) models among those evaluated in this work. The mitigation strategies are: (i) applying a Kalman filter to denoise the target object's state estimates, and (ii) retraining the models using a more realistic noise-augmented dataset. For the first strategy, a Kalman filter is applied to the target object's state estimates. The Kalman filter employs a constant-velocity motion model and its process (Q) and measurement (R) noise covariance matrices are configured as follows: Q = diag(0.05, 0.05, 0.10, 0.10) and R = diag ($0.30^2$, $0.30^2$, $0.50^2$, $0.50^2$). The covariance matrices were empirically selected by evaluating different configurations to improve denoising performance. For the second strategy, DeMo++ and Forecast-MAE are retrained on a modified version of the Argoverse 2 dataset in which the history of each target object (i.e., its position, speed, and heading) is perturbed according to the sensing and V2X noise models presented in Section III-C . For V2X noise, the standard deviation $\sigma_{\mathrm{V2X}}$ is randomly sampled from the [0.1, 2.0] m interval. As in their original implementations, both DeMo++ and Forecast-MAE are retrained using $T_{\mathrm{H}}$ = 5 seconds of the target object's history as input data and predict future trajectories over the next $T_{\mathrm{F}}$ = 6 seconds, with trajectories sampled at 10 Hz. The models' architectures, loss functions, and training configurations remain unchanged during retraining. Thus, the only modification to their original training process is the injection of synthetic sensing and V2X noise into the training data. We openly release the noise-augmented Argoverse 2 dataset used for the retraining to support further research on robust trajectory prediction [14].

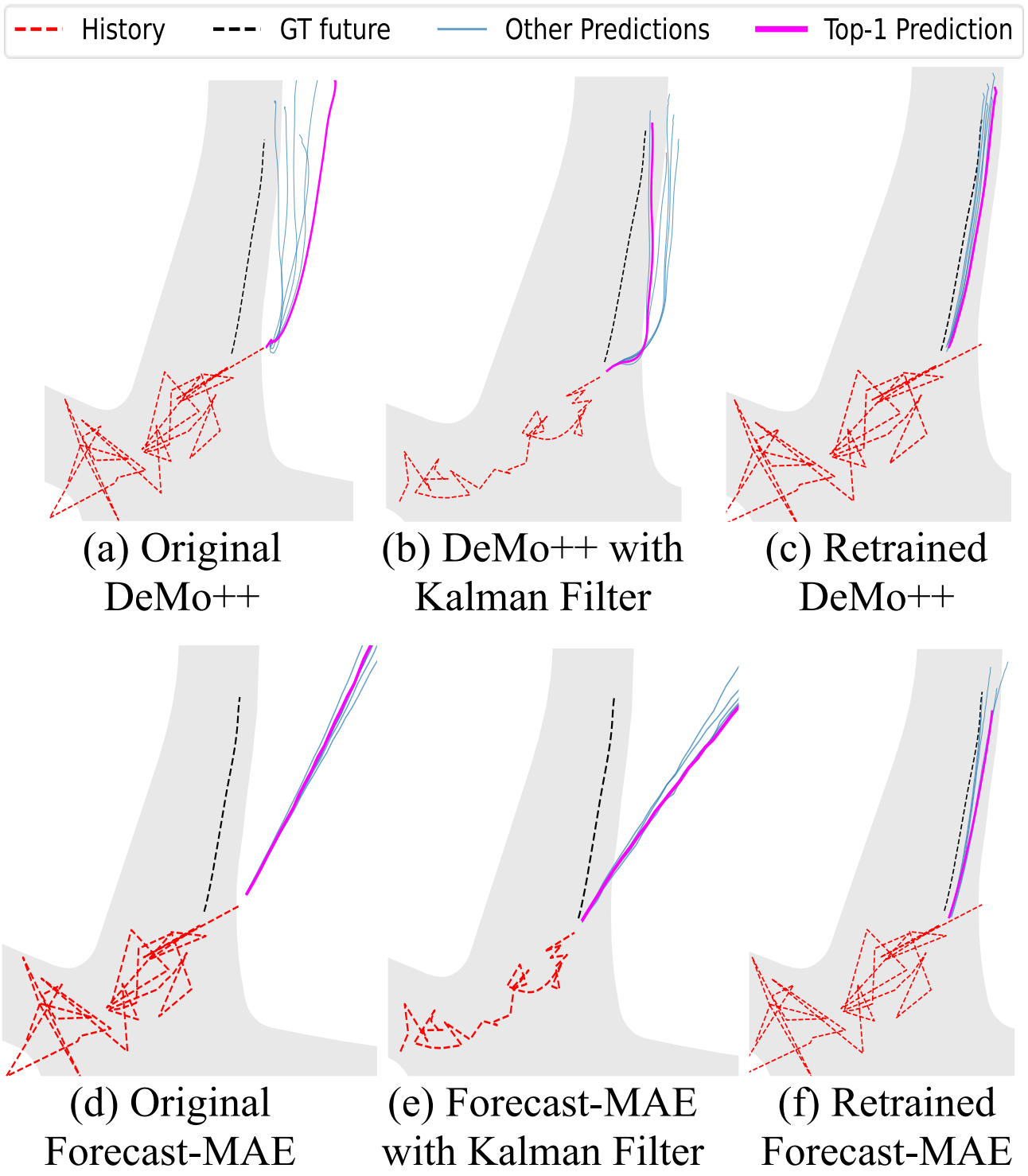

**Fig. 6.** Qualitative comparison of DeMo++ and Forecast-MAE trajectory predictions under extreme noise levels ($\sigma_{V2X}$ = 5.0 m) using a Kalman filter to denoise the data or retraining the models.

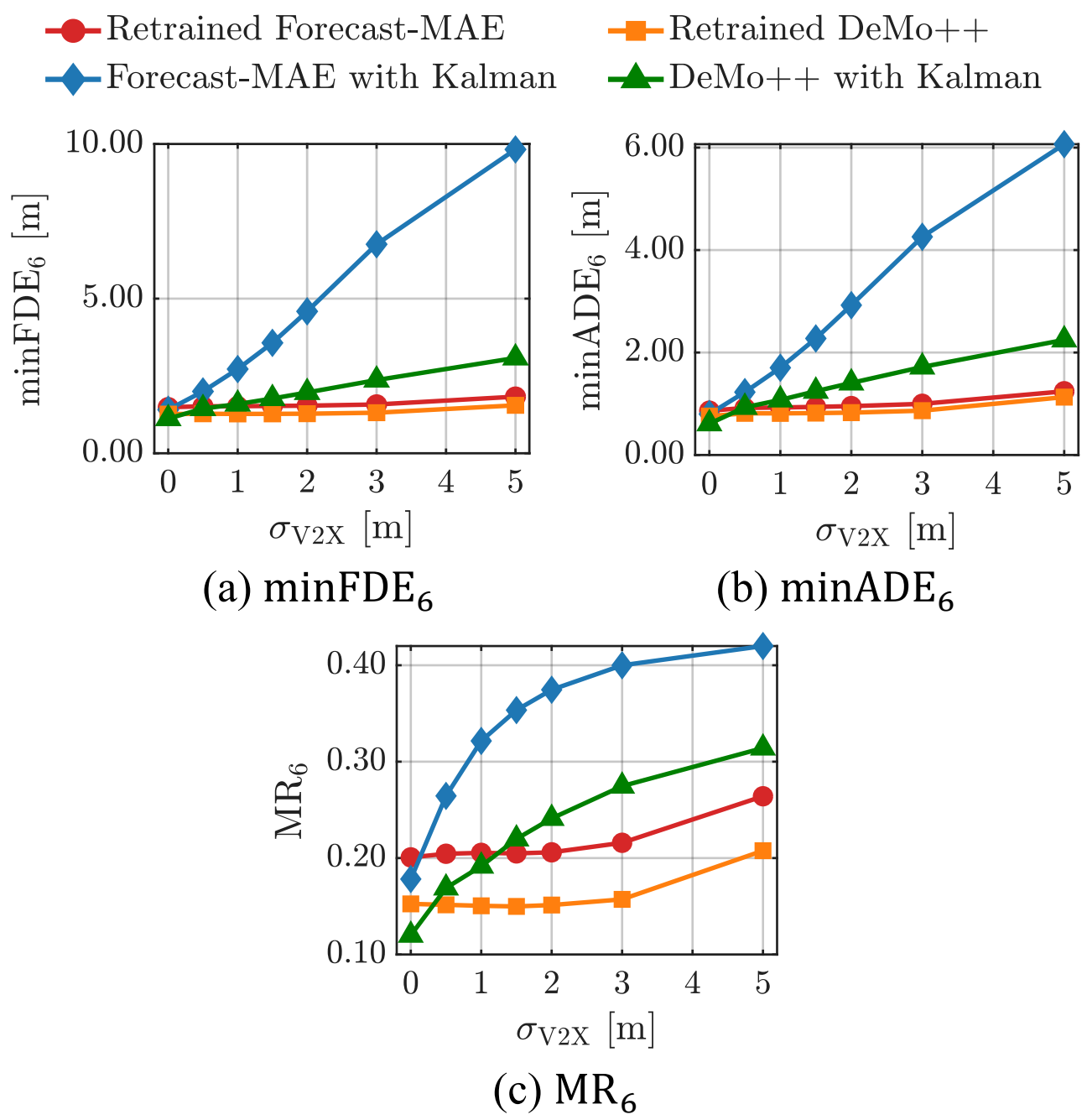

**Fig. 7.** Comparison of noise mitigation strategies under increasing V2X noise levels ($K$ = 6, sensing noise included).

Fig. 6 qualitatively compares the impact of Kalman filtering and noise-augmented retraining on the trajectory prediction accuracy of DeMo++ (Fig. 6(a)–(c)) and Forecast-MAE (Fig. 6(d)–(f)) under very high V2X noise conditions ($\sigma_{\mathrm{V2X}}$ = 5.0 m) to illustrate the robustness of the retrained models under challenging conditions. The figures use the same representative Argoverse 2 scenes previously considered for DeMo++ (Fig. 1(f)) and Forecast-MAE (Fig. 4(c)), respectively. Fig. 6(a) and Fig. 6(d) show the predictions of the original models without any mitigation strategy. Fig. 6(b) shows that Kalman filtering effectively denoises the target object's state estimates and partially improves DeMo++ trajectory prediction accuracy. However, the Kalman filter cannot completely remove the strong V2X noise, and DeMo++ predictions consequently remain misaligned with the ground-truth future trajectory. In contrast, Fig. 6(c) shows that retraining DeMo++ on noise-augmented data substantially mitigates the impact of noisy object state estimates. The retrained model outputs highly accurate and kinematically plausible predictions that closely align with the ground truth even under the strong V2X noise conditions ($\sigma_{\mathrm{V2X}}$ = 5.0 m). Fig. 6(e) and Fig. 6(f) show that similar trends are observed for Forecast-MAE, the most noise-sensitive model analyzed in this work. While Kalman filtering has a limited impact on trajectory prediction accuracy compared with the original model (Fig. 6(d)), retraining Forecast-MAE substantially improves its robustness to noisy object state estimates compared with both the original model with Kalman filtering (Fig. 6(e)) and the original model without Kalman filtering (Fig. 6(d)). These results suggest that noise-augmented retraining is more effective than input denoising alone for both models.

Fig. 7 quantitatively compares the impact of Kalman filtering and noise-augmented retraining on the trajectory prediction accuracy of DeMo++ and Forecast-MAE. The figure reports the $\mathrm{minFDE}_K$, $\mathrm{minADE}_K$, and $\mathrm{MR}_K$ metrics for the $K = 6$ setting under sensing noise and increasing V2X noise levels ranging from $\sigma_{\mathrm{V2X}} = 0.0$ m to $\sigma_{\mathrm{V2X}} = 5.0$ m. Results for $K = 1$ are not reported since they exhibit similar trends to those observed for K = 6. The results corroborate the qualitative observations in Fig. 6, showing that retraining is substantially more effective than denoising the input data at mitigating the impact of object state estimation uncertainties. For V2X noise levels included during retraining ($\sigma_{\mathrm{V2X}} \leq 2.0$ m), both retrained models maintain prediction accuracy close to that achieved under noise-free conditions, whereas Kalman filtering results in increasing prediction errors, particularly for Forecast-MAE. For V2X noise levels beyond those included in the retraining dataset ($\sigma_{\mathrm{V2X}} > 2.0$ m), retraining limits the degradation in prediction accuracy for both models, whereas the effectiveness of Kalman filtering is substantially reduced. The use of a Kalman filter only partially mitigates the models' sensitivity to high object state estimation uncertainties, with all three metrics greatly increasing as V2X noise intensifies, particularly for Forecast-MAE. Under very high V2X noise ($\sigma_{\mathrm{V2X}} = 5.0$ m), the $\mathrm{minFDE}_6$ achieved with Kalman filtering reaches approximately 3.0 m for DeMo++ and 9.8 m for Forecast-MAE, compared with approximately 1.5 m and 1.8 m for their respective retrained models.

Fig. 8 complements this analysis by reporting the relative increase in $\mathrm{minFDE}_6$, $\mathrm{minADE}_6$, and $\mathrm{MR}_6$ with respect to their reference values measured on the original noise-free dataset. For V2X noise levels included during training ($\sigma_{\mathrm{V2X}} \leq 2.0$ m), the retrained models exhibit a relative increase of less than approximately 1.2x across all three metrics, demonstrating their ability to maintain prediction accuracy close to that achieved under noise-free conditions. In contrast, Kalman filtering results in substantially higher relative increases for both models, reaching approximately 2.5x for DeMo++ and 4x for Forecast-MAE at $\sigma_{\mathrm{V2X}} = 2.0$ m. For V2X noise levels beyond those considered during retraining ($\sigma_{\mathrm{V2X}} > 2.0$ m), the relative increase in all metrics remains below approximately 1.5x for the retrained models, even at $\sigma_{\mathrm{V2X}} = 5.0$ m, whereas with Kalman filtering, the relative increase reaches approximately 3x for DeMo++ and 8x for Forecast-MAE. By comparison, the original DeMo++ trained on the noise-free Argoverse 2 dataset (see Fig. 2(b)) exhibits a relative increase that can be as large as 4x when evaluated under a V2X noise level of $\sigma_{\mathrm{V2X}} = 2.0$ m. These results highlight the importance of training trajectory prediction models on datasets that account for real-world object state estimation uncertainties. Exposing the models to realistic sensing and V2X noise during the training phase improves robustness to noisy inputs and helps limit the degradation in prediction accuracy even at noise levels beyond those encountered during training.

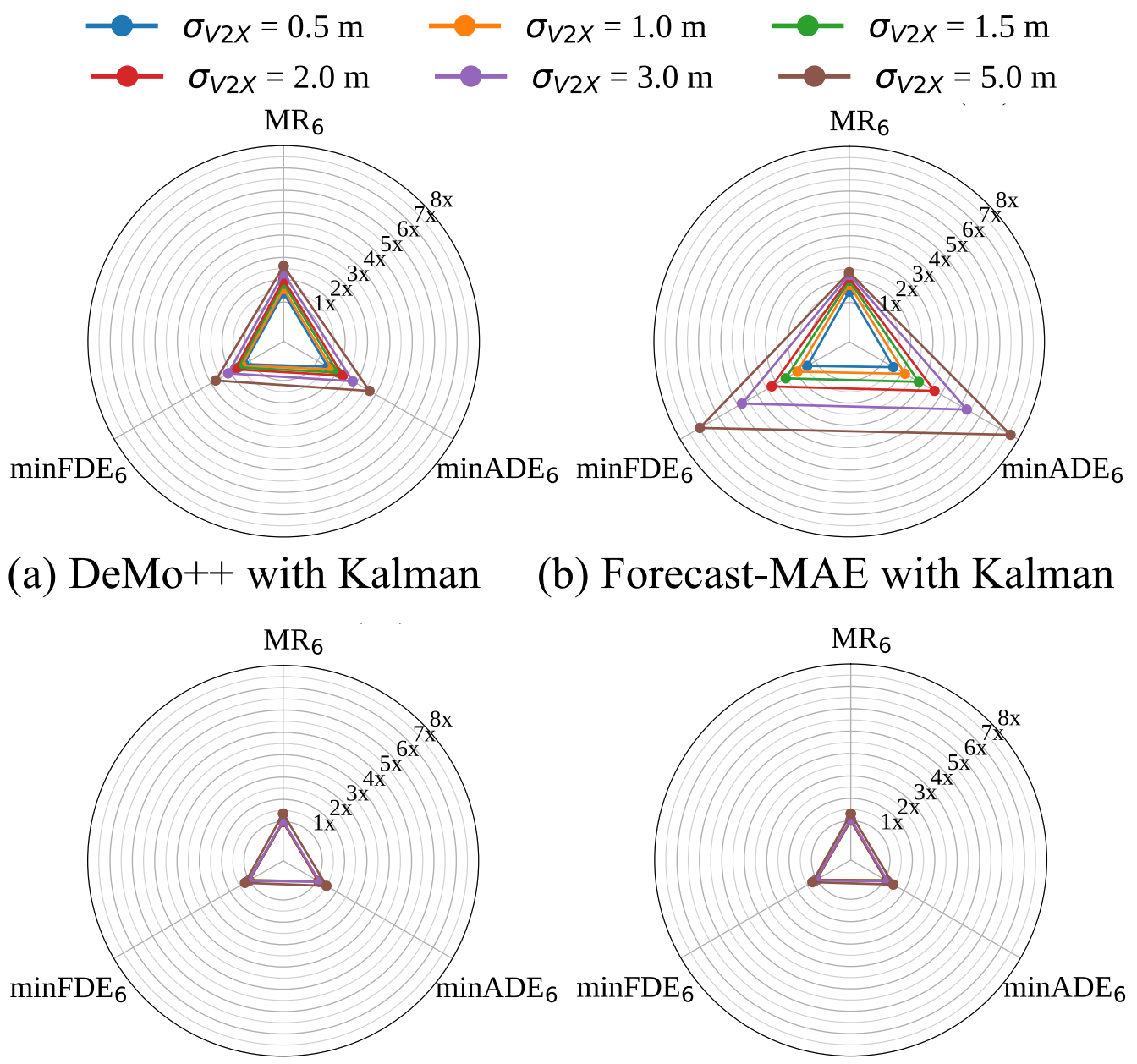

(a) DeMo++ with Kalman (b) Forecast-MAE with Kalman

(c) Retrained DeMo++ (d) Retrained Forecast-MAE

**Fig. 8.** Comparison of mitigation strategies. Relative metrics degradation under increasing $N_{V2X}$ noise ($K = 6$).

## VI. Conclusions

State-of-the-art Transformer-based interaction-aware trajectory prediction models are commonly trained and evaluated on highly curated noise-free datasets that do not reflect the object state estimation uncertainties typically encountered in real-world deployments. In this work, we analyzed the sensitivity of five state-of-the-art trajectory prediction models to sensing- and V2X-related object state estimation uncertainties present in connected and automated mobility. Our results show that, while the considered models are largely robust to sensing noise, their prediction accuracy substantially degrades under the higher noise levels associated with object state information received via V2X communications. To mitigate this degradation, we numerically demonstrated that retraining trajectory prediction models on noise-augmented datasets is a more effective strategy than trying to denoise the input data using a Kalman filter. Retrained models largely eliminate any accuracy degradation under noise conditions similar to those included in the noise-augmented retraining dataset. For noise levels beyond those encountered during training, retrained models reduce the accuracy degradation from a factor of 4 to approximately 1.5 compared with the original models trained on noise-free datasets. Our results underscore the importance of training and evaluating trajectory prediction models using datasets that faithfully reflect real-world object state estimation uncertainties. The strong robustness achieved through retraining also raises questions about the need for additional denoising modules or custom AI architectures designed to handle noisy inputs (e.g., [9] and [24]), which introduce additional computational overhead and inference latency. To facilitate further research, we publicly release the noise-augmented version of the Argoverse 2 dataset generated and used in this work at [14].